\documentclass{article}
\usepackage{spconf,amsmath,graphicx}

\usepackage{enumitem}
\setlist{nosep, leftmargin=14pt}

\usepackage[table,xcdraw]{xcolor} 
\usepackage{amsfonts}  
\usepackage{bm}
\usepackage{array}     
\usepackage{booktabs}  
\usepackage{multirow}  
\usepackage{makecell}  
\usepackage{subcaption}
\usepackage{pifont}    
\usepackage{url}
\usepackage[hidelinks]{hyperref}  
\usepackage{pgfplots}
\pgfplotsset{compat=1.18}
\usepgfplotslibrary{groupplots}
\usetikzlibrary{matrix,shapes.geometric,calc}

\definecolor{colOurs}{HTML}{378ADD}      
\definecolor{colBaseline}{HTML}{D85A30}  
\definecolor{colRandom}{HTML}{762A83}    
\definecolor{colLoco}{HTML}{009E73}      
\colorlet{colOursFull}{colOurs}          
\colorlet{colNtssl}{colBaseline}         
\newcommand{\mplfont}{\fontfamily{DejaVuSans-TLF}\selectfont}

\newcommand{\rc}[1]{REAL-Colon}
\newcommand{\sota}[1]{state-of-the-art}
\newcommand{\SotA}[1]{State-of-the-Art}
\newcommand{\etal}[1]{\textit{et al.}}
\newcommand{\cmark}{\ding{51}}%
\newcommand{\xmark}{\ding{55}}%

\title{Annotation-Free Representation Learning for Polyp Retrieval in Colonoscopy Videos}
\name{Luca Parolari$^{1}$ \qquad Lamberto Ballan$^{1}$ \qquad Carlo Biffi$^{2,\dagger}$ \qquad Loic Le Folgoc$^{3,\dagger}$}
\address{$^{1}$ Department of Mathematics, University of Padova, Padova, Italy \\
    $^{2}$ Cosmo Intelligent Medical Devices, Dublin, Ireland \\
    $^{3}$ LTCI, Telecom Paris, Institut Polytechnique de Paris, Palaiseau, France}
\begin{document}
%
\maketitle
{\renewcommand{\thefootnote}{\fnsymbol{footnote}}\footnotetext[2]{Shared senior authorship.}}
\begin{abstract}
Colonoscopy is inherently comparative: reliable visual assessment of polyps often depends on complementary views acquired under different viewpoints and imaging conditions.
Retrieving such views requires polyp tracklet representations, which recent contrastive methods learn from temporal cues but still build from manually annotated bounding boxes.
We propose an annotation-free framework that constructs tracklet views directly from raw colonoscopy videos, enabling temporal self-supervised learning on full-procedure videos, including frames without visible polyps.
On \rc{}, our method achieves state-of-the-art results on tracklet retrieval and re-identification, outperforming both self-supervised and supervised baselines.
We then introduce few-shot polyp retrieval, a task closer to clinical use, where a few observations of a target polyp are used to retrieve its remaining tracklets: with only two examples, our method surpasses most competing methods given ten.
Finally, we show that a modest number of unlabeled procedures suffices for competitive performance, making on-site model development more practical.
Code is available at {\urlstyle{rm}\url{https://github.com/lparolari/freerep}}.
\end{abstract}
\section{Introduction} 

Colonoscopy is the gold standard for the prevention and early detection of colorectal cancer~\cite{hassan2021performance}. Deep learning increasingly supports it through computer-aided polyp detection~\cite{dominitz2026impact} and characterization~\cite{hassan2024computer}.
Polyp retrieval aims to recover additional views of a target lesion from colonoscopy videos, starting from one or a few observations.
Aggregating information across these complementary views could support polyp characterization and facilitate lesion assessment and reporting during post-procedure review~\cite{naito2025usefulness,DBLP:conf/icmcs/ChenCCYQX23}.
However, learning representations of polyp tracklets---sequences of consecutive detections of the same polyp---is challenging: different polyps may look similar, while the same polyp may vary substantially in appearance over time~\cite{gussman2026assessing}.
Representations must therefore be robust to appearance changes while remaining discriminative across polyps.

Supervised contrastive learning on \emph{polyp entity annotations}, \textit{i.e.}, labels linking observations of the same polyp, yields strong tracklet representations~\cite{parolari2025temporally}. Yet, such labels are costly to collect, as they require expert clinicians.
Temporal self-supervised methods reduce this cost by deriving positive pairs from temporal continuity: by splitting tracklets into temporal segments~\cite{intrator2023self,parolari2025towards}, by integrating textual reports through momentum contrast~\cite{xiang2024vtreid}, or by sampling pseudo-positives among temporally close tracklets of the same procedure~\cite{parolari2026contrastive}.

These approaches nevertheless rely on ground-truth bounding boxes to construct polyp-centered training inputs, providing explicit spatial supervision during representation learning.
Such annotations are costly and time-consuming to obtain.
They also restrict training to annotated polyp observations, excluding the large majority of each procedure, namely frames without visible polyps, which still carry useful information about normal colonic appearance.

In this work, we learn polyp tracklet representations directly from raw colonoscopy videos, without bounding boxes or polyp identity labels.
Since tracklets can no longer be polyp-centered, we redefine them as sequences of consecutive frames. However, our experiments show that global frame-level views underperform because of the lack of locality. Therefore, we adopt a temporally consistent random cropping strategy that restores locality without annotations.
This enables temporal self-supervision on full-procedure videos and yields \sota{} results on \rc{} for tracklet retrieval and re-identification.
We then introduce \emph{few-shot polyp retrieval}, which retrieves the remaining tracklets of a lesion from a few examples, as in post-procedure review.
With only two support tracklets, our method outperforms most baselines that use ten.

Our contributions are:
(i) an annotation-free framework that achieves \sota{} results by learning directly from raw full-procedure videos;
(ii) a novel few-shot polyp retrieval protocol;
(iii) analyses of the trade-off between spatial supervision and data scale, and of cross-cohort robustness.

\section{Annotation-Free Representation Learning Framework}
\label{sec:method}

We train a visual encoder by contrastive learning without the two sources of supervision of polyp-centric settings: ground-truth polyp detections and polyp entity annotations.
Temporally consistent crops of raw frames recover the locality of polyp-centered views (Sec.~\ref{sec:annotation_free}), extending training to full-procedure videos, including frames without visible polyps (Sec.~\ref{sec:full_procedure}).
Pseudo-positive associations derived from temporal proximity within each video replace polyp entity labels (Sec.~\ref{sec:temporal_ssl}), enabling annotation-free representation learning.

\subsection{Annotation-Free Tracklet Views}
\label{sec:annotation_free}

Polyp-centric frameworks build each training sample from ground-truth polyp detections.
Without them, we redefine a tracklet as a short sequence of consecutive frames from the same video, whether or not a polyp is visible.

The most direct view is then a sequence of full frames, which removes the spatial prior that focused the encoder on the polyp. However, this full-frame baseline loses 15\% (relative) with respect to the crop-based \sota{} trained with ground-truth detections.
Full frames shift the visual domain toward global endoscopic views with large background regions, instruments, specular highlights, and blur.

We therefore keep local views but sample them without detections: a random crop is drawn once per tracklet and applied to all its frames, preserving spatial coherence.
This yields a 27\% relative improvement over full frames, surpasses the crop-based \sota{} by 8\%, and performs comparably to a synthetic box crop based on dataset-level bounding-box statistics, which, unlike ours, requires annotations (Tab.~\ref{tab:ablation}).

\subsection{Full-Procedure Training}
\label{sec:full_procedure}

Since views no longer depend on detections, training is not confined to annotated polyp tracklets and extends to raw full-procedure videos, including the mucosal patterns, lumen views, and other colonoscopy-specific content that polyp-centric methods discard.

We build the training set from \rc{}, which contains 60 full-procedure videos from four acquisition cohorts.
We follow the split of~\cite{parolari2026contrastive}, but retain the full content of its 27 training videos with polyps and add the 14 videos without visible polyps that it omitted.
The resulting 41-video training set, a superset of the previously used annotated data, contains 1,725,537 frames, of which only 171,957 (approximately 10\%) show at least one polyp.

\subsection{Temporal Self-Supervision}
\label{sec:temporal_ssl}

We optimize the tracklet encoder with the temporal self-supervised framework of~\cite{parolari2026contrastive}, which derives pseudo-positive associations from the temporal structure of colonoscopy videos rather than from polyp identity labels.
We leave the objective unchanged and modify only its input domain: tracklets are no longer required to contain a polyp, which allows training on full-procedure videos.

Formally, each training tracklet $x_i$ has a source video $v_i$ and a temporal position $p_i$.
For an anchor $a_i$, the candidate pseudo-positives are the tracklets of the same video ($v_j=v_i$), ranked by temporal distance.
We sample $\kappa$ of them without replacement from an exponential distribution over ranks with temperature $\tau$, obtaining a bag $B_i=\{b_{ik}\}_{k=1}^{\kappa}$; a curriculum increases $\tau$ in training, from close to broader context.

We minimize the noise-aware contrastive loss~\cite{parolari2026contrastive}:
\[
\mathcal{L}
=
-\sum_i
\log
\frac{\sum_k e^{s(z_{ik},y_i)}}
{\sum_k e^{s(z_{ik},y_i)} + \sum_{j\neq i} \sum_k e^{s(z_{ik},y_j)}},
\]
where $y_i=f_\theta(a_i)$ and $z_{ik}=f_\theta(b_{ik})$ are the encoder representations of the anchor and of the $k$-th element of its bag, and $s(\cdot,\cdot)$ is a similarity function.
Summing over the bag lets the model weight the most relevant pseudo-positives instead of forcing all temporal neighbors to match the anchor equally.

\section{Experimental Results}
\label{sec:experiments}

\begin{table*}[t]
\caption{
Comparison with \SotA{}: our approach outperforms self-supervised, weakly supervised, and fully supervised baselines.
\textbf{Bold}: best; \underline{underline}: second best.
}
\centering
\footnotesize
\begin{tabular}{lcc|cc|cc|ccccc}
\toprule

\multirow{3}{*}{Method}
& &
& \multicolumn{2}{c|}{\textit{Tracklet Retrieval}} 
& \multicolumn{2}{c|}{\textit{Tracklet ReID}}
& \multicolumn{5}{c}{\textit{Few-Shot Polyp Retrieval}}
\\

& Params & FLOPs 
& mAP & HR@1
& AUROC & AUPR 
& \multicolumn{5}{c}{mAP ($\uparrow$)}
\\

& (M) & (G) 
& $(\uparrow)$ & $(\uparrow)$
& $(\uparrow)$ & $(\uparrow)$ 
& $K{=}1$ & $K{=}2$ & $K{=}3$ & $K{=}5$ & $K{=}10$
\\

\midrule

\multicolumn{3}{l|}{\textit{General Purpose FM}} & & & & & & & & & \\

DinoV2-S/14~\cite{oquab2024dinov2}
& 22.1 & 44.2
& 47.31 & 89.81
& 80.07 & 29.77
& 47.91 & 57.04 & 62.73 & 65.64 & 72.57
\\

DinoV2-g/14~\cite{oquab2024dinov2}
& 1136 & 2331
& 48.29 & 90.35
& 78.23 & 27.04
& 48.08 & 57.20 & 62.40 & 66.46 & 72.13
\\

DinoV3-S/16~\cite{simeoni2025dinov3}
& 21.6 & 34.7
& 48.95 & 88.99
& 80.47 & 31.77 
& 47.60 & 57.36 & 62.95 & 66.80 & 76.59
\\

V-JEPA2~\cite{assran2025vjepa2}
& 326.0 & 272.1
& 29.01 & 70.79
& 69.84 & 17.14 
& 22.29 & 28.46 & 32.70 & 33.38 & 39.93
\\

\hline
\multicolumn{3}{l|}{\textit{Endoscopy FM}} & & & & & & & & & \\

EndoFM~\cite{wang2023endofm}
& 121.3 & 190.1
& 45.75 & 87.64
& 79.71 & 31.91 
& 45.12 & 56.28 & 61.80 & 65.46 & 73.03
\\

EndoFM-LV~\cite{wang2025endofmlv}
& 121.3 & 190.1
& 28.21 & 63.18
& 67.56 & 19.28 
& 22.54 & 28.22 & 28.92 & 30.95 & 38.22
\\

EndoViT~\cite{batic2024endovit}
& 86.6 & 134.9
& 31.12 & 68.34
& 68.26 & 17.51 
& 28.18 & 35.60 & 38.68 & 40.77 & 47.77 
\\

SurgeNet~\cite{jaspers2026surgenet}
& 24.3 & 31.0
& 44.75 & 85.87
& 75.86 & 25.23 
& 47.14 & 58.25 & 63.37 & 68.18 & 75.87
\\

\hline
\multicolumn{3}{l|}{\textit{Fully supervised}} & & & & & & & & & \\

Sup~\cite{parolari2025temporally}
& 25.6 & 33.1
& 54.72 & 81.93
& 93.85 & 41.58 
& 50.60 & 55.75 & 58.23 & 59.60 & 63.22
\\

TA~\cite{parolari2025temporally} 
& 25.6 & 33.1
& 57.52 & 84.24
& \underline{94.33} & 43.81 
& 53.12 & 57.94 & 60.49 & 62.33 & 68.52 
\\

\hline
\multicolumn{3}{l|}{\textit{Weakly supervised}} & & & & & & & & & \\

MVE~\cite{intrator2023self}
& 25.6 & 33.1
& 42.05 & 81.25
& 79.73 & 27.47
& 48.41 & 56.25 & 58.47 & 60.58 & 65.89 
\\

TPC~\cite{parolari2025towards}
& 25.6 & 33.1
& 43.25 & 75.82
& 78.40 & 27.58
& 51.15 & 59.46 & 63.00 & 65.51 & 69.90 
\\

NTSSL~\cite{parolari2026contrastive}
& 25.6 & 33.1
& \underline{61.49} & \underline{91.17}
& 93.89 & \underline{51.78} 
& \underline{59.24} & \underline{67.64} & \underline{71.36} & \underline{74.69} & \underline{81.34}
\\

\hline
\multicolumn{3}{l|}{\textit{Self-supervised}} & & & & & & & & & \\

MVE~\cite{intrator2023self}
& 25.6 & 33.1
& 46.22 & 80.43
& 89.74 & 31.93
& 44.69 & 51.67 & 54.03 & 54.62 & 59.30
\\

TPC~\cite{parolari2025towards}
& 25.6 & 33.1
& 49.94 & 84.65
& 91.60 & 35.66
& 46.96 & 54.73 & 57.49 & 58.79 & 62.83
\\

\rowcolor{orange!20}
Ours
& 25.6 & 33.1
& \textbf{66.47} & \textbf{92.80}
& \textbf{95.03} & \textbf{54.91} 
& \textbf{63.46} & \textbf{73.10} & \textbf{75.58} & \textbf{78.80} & \textbf{85.72} 
\\

\bottomrule
\end{tabular}
\label{tab:sota}
\end{table*}

\subsection{Dataset and Downstream Setup}
\label{sec:dataset}

We use \rc{}~\cite{biffi2024real}, which contains 60 full-procedure colonoscopy videos from four acquisition cohorts recorded at six clinical centers on three continents with equipment from two manufacturers. We follow the setup of~\cite{parolari2026contrastive}.
Training uses 41 videos without any annotation (Sec.~\ref{sec:full_procedure}); evaluation uses the 19-video test set (47 polyps), which contains only tracklets with visible polyps.

\noindent\textbf{Tracklet retrieval and ReID}\quad
Following~\cite{parolari2026contrastive}, each test tracklet is used in turn as a query, and the remaining tracklets are ranked by the cosine similarity of their embeddings to it; we report mAP and HR@1, the fraction of queries whose top-ranked tracklet depicts the same polyp.
For re-identification, the cosine similarity of each tracklet pair is thresholded to separate same-polyp from different-polyp pairs, and performance is summarized by AUROC and AUPR.

\noindent\textbf{Few-shot polyp retrieval}\quad
Tracklet retrieval treats each tracklet as an independent query, whereas in clinical review an expert may hold one or a few observations of a polyp and search for its other appearances.
We therefore retrieve at the polyp level: for each target polyp, we randomly sample a support set $S=\{s_1,\ldots,s_K\}$ of $K\in\{1,2,3,5,10\}$ of its tracklets and rank each gallery tracklet $g\in\mathcal{G}$ by $\mathrm{sim}(g,S)=\max_{s_i\in S}\cos(f_\theta(g),f_\theta(s_i))$, keeping the support tracklets as separate exemplars.
Random sampling models an average review scenario rather than a curated selection; we repeat it $R=50$ times and report the mean mAP.

\subsection{Implementation Details}
The tracklet encoder follows prior work~\cite{intrator2023self,parolari2025towards,parolari2025temporally,parolari2026contrastive}: an ImageNet-pretrained ResNet-50~\cite{he2016deep} extracts frame features, which a three-layer transformer encoder (8 heads, $d_{\mathrm{model}}=256$) aggregates and a two-layer MLP ($256\rightarrow128$) projects for contrastive learning.
We use $\kappa=4$ and a batch size of 60. The curriculum schedules $\tau$ in the exponential distribution from $\tau_{\min}=0.3$ to $\tau_{\max}=12$, using a cosine scheduler~\cite{parolari2026contrastive}.
Videos are temporally subsampled by a factor of four, and a tracklet spans $L=8$ subsampled frames (about one second); views are random resized crops (scale range $(0.08,1)$, $232\times232$ pixels) sampled once per tracklet, with no further augmentation.
We train for 25 epochs with AdamW (learning rate $10^{-5}$, weight decay $10^{-4}$) on a single NVIDIA A100 GPU.
At test time, as in~\cite{parolari2025towards,parolari2025temporally,parolari2026contrastive}, tracklets are built by linking ground-truth detections with an IoU of at least $0.1$, each enlarged to $5\times$ the bounding-box diagonal.

\subsection{Comparison with \SotA{}}
\label{sec:exp_sota}

Tab.~\ref{tab:sota} compares our approach with the \sota{}.
Self-supervised baselines are re-trained on our training set, while weakly and fully supervised baselines use its annotated subset.
Our framework outperforms previous self-supervised approaches by 33.10\% in retrieval mAP and by 3.74\% and 53.98\% in ReID AUROC and AUPR (relative). MVE~\cite{intrator2023self} and TPC~\cite{parolari2025towards} draw positive views from distinct segments of a tracklet, presuming a polyp that stays visible throughout, which rarely holds for generic frame sequences.

Our method also surpasses NTSSL~\cite{parolari2026contrastive}, a weakly supervised method that shares our objective but relies on bounding boxes, and the fully supervised baselines on all metrics; against TA~\cite{parolari2025temporally}, our method slightly improves AUROC and improves AUPR by 25.34\%, \textit{i.e.}, with fewer false positives.
Foundation models give strong generic features but are not optimized for the temporal, low-semantics structure of colonoscopy; endoscopy-specific ones rarely beat general-purpose ones, consistent with their smaller scale and surgical training data.

\noindent\textbf{Few-shot polyp retrieval}\quad
All methods benefit from additional support observations, confirming that single views of a polyp are ambiguous.
Moving from $K=1$ to $K=2$ raises our mAP from 63.46 to 73.10, which already exceeds the $K=10$ performance of nearly all baselines; the only exceptions are DinoV3-S/16, SurgeNet, and NTSSL, which uses ground-truth boxes during training.
With $K=10$, our method reaches 85.72 mAP.

\subsection{Supervision vs. Scaling}
\label{sec:exp_scaling}

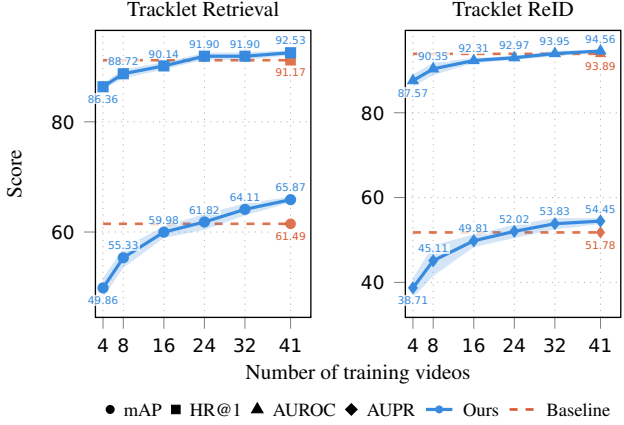
\begin{figure}[t]
    \centering
\begin{tikzpicture}

\begin{groupplot}[
    group style={group size=2 by 1, horizontal sep=1.25cm},
    scale only axis,
    width=2.85cm,
    height=3.8cm,
    title style={font=\footnotesize, yshift=-1.5mm},
    xmin=2.5, xmax=45,
    xtick={4,8,16,24,32,41},
    label style={font=\footnotesize},
    tick label style={font=\scriptsize\mplfont},
    /pgf/number format/assume math mode=true,
    yticklabel style={text width=1.3em, align=right},
    tick align=outside, xtick pos=bottom, ytick pos=left,
    grid=major, grid style={dotted, gray!60},
    axis line style={line width=0.5pt},
    enlarge y limits=0.08,
    clip=false,
]

\nextgroupplot[title={Tracklet Retrieval}, ylabel={Score}]

\addplot[colBaseline, line width=1pt, dashed, opacity=0.85,
    mark=*, mark size=1.5pt, mark options={solid}, mark indices={2}]
    coordinates {(4,61.49) (41,61.49)};
\node[font=\fontsize{4}{4.5}\selectfont\mplfont, colBaseline, inner sep=0.4pt, fill=white, fill opacity=0.75, text opacity=1, anchor=north, yshift=-2.5pt]
    at (axis cs:41,61.49) {61.49};
\addplot[draw=none, fill=colOurs, fill opacity=0.22]
    coordinates {(4,51.62) (8,57.15) (16,61.03) (24,63.26) (32,65.21) (41,66.38) (41,65.36) (32,63.01) (24,60.38) (16,58.93) (8,53.51) (4,48.10)} -- cycle;
\addplot[colOurs, line width=1.2pt, mark=*, mark size=1.5pt]
    coordinates {(4,49.86) (8,55.33) (16,59.98) (24,61.82) (32,64.11) (41,65.87)};
\node[font=\fontsize{4}{4.5}\selectfont\mplfont, colOurs, inner sep=0.4pt, fill=white, fill opacity=0.75, text opacity=1, anchor=north, yshift=-2.5pt] at (axis cs:4,49.86) {49.86};
\node[font=\fontsize{4}{4.5}\selectfont\mplfont, colOurs, inner sep=0.4pt, fill=white, fill opacity=0.75, text opacity=1, anchor=south, yshift=2.5pt] at (axis cs:8,55.33) {55.33};
\node[font=\fontsize{4}{4.5}\selectfont\mplfont, colOurs, inner sep=0.4pt, fill=white, fill opacity=0.75, text opacity=1, anchor=south, yshift=2.5pt] at (axis cs:16,59.98) {59.98};
\node[font=\fontsize{4}{4.5}\selectfont\mplfont, colOurs, inner sep=0.4pt, fill=white, fill opacity=0.75, text opacity=1, anchor=south, yshift=2.5pt] at (axis cs:24,61.82) {61.82};
\node[font=\fontsize{4}{4.5}\selectfont\mplfont, colOurs, inner sep=0.4pt, fill=white, fill opacity=0.75, text opacity=1, anchor=south, yshift=2.5pt] at (axis cs:32,64.11) {64.11};
\node[font=\fontsize{4}{4.5}\selectfont\mplfont, colOurs, inner sep=0.4pt, fill=white, fill opacity=0.75, text opacity=1, anchor=south, yshift=2.5pt] at (axis cs:41,65.87) {65.87};

\addplot[colBaseline, line width=1pt, dashed, opacity=0.85,
    mark=square*, mark size=1.5pt, mark options={solid}, mark indices={2}]
    coordinates {(4,91.17) (41,91.17)};
\node[font=\fontsize{4}{4.5}\selectfont\mplfont, colBaseline, inner sep=0.4pt, fill=white, fill opacity=0.75, text opacity=1, anchor=north, yshift=-2.5pt]
    at (axis cs:41,91.17) {91.17};
\addplot[draw=none, fill=colOurs, fill opacity=0.22]
    coordinates {(4,86.91) (8,89.55) (16,90.80) (24,92.52) (32,92.60) (41,92.99) (41,92.07) (32,91.20) (24,91.28) (16,89.48) (8,87.89) (4,85.81)} -- cycle;
\addplot[colOurs, line width=1.2pt, mark=square*, mark size=1.5pt]
    coordinates {(4,86.36) (8,88.72) (16,90.14) (24,91.90) (32,91.90) (41,92.53)};
\node[font=\fontsize{4}{4.5}\selectfont\mplfont, colOurs, inner sep=0.4pt, fill=white, fill opacity=0.75, text opacity=1, anchor=north, yshift=-2.5pt] at (axis cs:4,86.36) {86.36};
\node[font=\fontsize{4}{4.5}\selectfont\mplfont, colOurs, inner sep=0.4pt, fill=white, fill opacity=0.75, text opacity=1, anchor=south, yshift=2.5pt] at (axis cs:8,88.72) {88.72};
\node[font=\fontsize{4}{4.5}\selectfont\mplfont, colOurs, inner sep=0.4pt, fill=white, fill opacity=0.75, text opacity=1, anchor=south, yshift=2.5pt] at (axis cs:16,90.14) {90.14};
\node[font=\fontsize{4}{4.5}\selectfont\mplfont, colOurs, inner sep=0.4pt, fill=white, fill opacity=0.75, text opacity=1, anchor=south, yshift=2.5pt] at (axis cs:24,91.90) {91.90};
\node[font=\fontsize{4}{4.5}\selectfont\mplfont, colOurs, inner sep=0.4pt, fill=white, fill opacity=0.75, text opacity=1, anchor=south, yshift=2.5pt] at (axis cs:32,91.90) {91.90};
\node[font=\fontsize{4}{4.5}\selectfont\mplfont, colOurs, inner sep=0.4pt, fill=white, fill opacity=0.75, text opacity=1, anchor=south, yshift=2.5pt] at (axis cs:41,92.53) {92.53};

\nextgroupplot[title={Tracklet ReID}]

\addplot[colBaseline, line width=1pt, dashed, opacity=0.85,
    mark=triangle*, mark size=1.5pt, mark options={solid}, mark indices={2}]
    coordinates {(4,93.89) (41,93.89)};
\node[font=\fontsize{4}{4.5}\selectfont\mplfont, colBaseline, inner sep=0.4pt, fill=white, fill opacity=0.75, text opacity=1, anchor=north, yshift=-2.5pt]
    at (axis cs:41,93.89) {93.89};
\addplot[draw=none, fill=colOurs, fill opacity=0.22]
    coordinates {(4,89.02) (8,91.80) (16,92.94) (24,93.44) (32,94.62) (41,94.75) (41,94.37) (32,93.28) (24,92.50) (16,91.68) (8,88.90) (4,86.12)} -- cycle;
\addplot[colOurs, line width=1.2pt, mark=triangle*, mark size=1.5pt]
    coordinates {(4,87.57) (8,90.35) (16,92.31) (24,92.97) (32,93.95) (41,94.56)};
\node[font=\fontsize{4}{4.5}\selectfont\mplfont, colOurs, inner sep=0.4pt, fill=white, fill opacity=0.75, text opacity=1, anchor=north, yshift=-2.5pt] at (axis cs:4,87.57) {87.57};
\node[font=\fontsize{4}{4.5}\selectfont\mplfont, colOurs, inner sep=0.4pt, fill=white, fill opacity=0.75, text opacity=1, anchor=south, yshift=2.5pt] at (axis cs:8,90.35) {90.35};
\node[font=\fontsize{4}{4.5}\selectfont\mplfont, colOurs, inner sep=0.4pt, fill=white, fill opacity=0.75, text opacity=1, anchor=south, yshift=2.5pt] at (axis cs:16,92.31) {92.31};
\node[font=\fontsize{4}{4.5}\selectfont\mplfont, colOurs, inner sep=0.4pt, fill=white, fill opacity=0.75, text opacity=1, anchor=south, yshift=2.5pt] at (axis cs:24,92.97) {92.97};
\node[font=\fontsize{4}{4.5}\selectfont\mplfont, colOurs, inner sep=0.4pt, fill=white, fill opacity=0.75, text opacity=1, anchor=south, yshift=2.5pt] at (axis cs:32,93.95) {93.95};
\node[font=\fontsize{4}{4.5}\selectfont\mplfont, colOurs, inner sep=0.4pt, fill=white, fill opacity=0.75, text opacity=1, anchor=south, yshift=2.5pt] at (axis cs:41,94.56) {94.56};

\addplot[colBaseline, line width=1pt, dashed, opacity=0.85,
    mark=diamond*, mark size=1.5pt, mark options={solid}, mark indices={2}]
    coordinates {(4,51.78) (41,51.78)};
\node[font=\fontsize{4}{4.5}\selectfont\mplfont, colBaseline, inner sep=0.4pt, fill=white, fill opacity=0.75, text opacity=1, anchor=north, yshift=-2.5pt]
    at (axis cs:41,51.78) {51.78};
\addplot[draw=none, fill=colOurs, fill opacity=0.22]
    coordinates {(4,41.01) (8,48.73) (16,51.30) (24,53.58) (32,55.16) (41,55.16) (41,53.74) (32,52.50) (24,50.46) (16,48.32) (8,41.49) (4,36.41)} -- cycle;
\addplot[colOurs, line width=1.2pt, mark=diamond*, mark size=1.5pt]
    coordinates {(4,38.71) (8,45.11) (16,49.81) (24,52.02) (32,53.83) (41,54.45)};
\node[font=\fontsize{4}{4.5}\selectfont\mplfont, colOurs, inner sep=0.4pt, fill=white, fill opacity=0.75, text opacity=1, anchor=north, yshift=-2.5pt] at (axis cs:4,38.71) {38.71};
\node[font=\fontsize{4}{4.5}\selectfont\mplfont, colOurs, inner sep=0.4pt, fill=white, fill opacity=0.75, text opacity=1, anchor=south, yshift=2.5pt] at (axis cs:8,45.11) {45.11};
\node[font=\fontsize{4}{4.5}\selectfont\mplfont, colOurs, inner sep=0.4pt, fill=white, fill opacity=0.75, text opacity=1, anchor=south, yshift=2.5pt] at (axis cs:16,49.81) {49.81};
\node[font=\fontsize{4}{4.5}\selectfont\mplfont, colOurs, inner sep=0.4pt, fill=white, fill opacity=0.75, text opacity=1, anchor=south, yshift=2.5pt] at (axis cs:24,52.02) {52.02};
\node[font=\fontsize{4}{4.5}\selectfont\mplfont, colOurs, inner sep=0.4pt, fill=white, fill opacity=0.75, text opacity=1, anchor=south, yshift=2.5pt] at (axis cs:32,53.83) {53.83};
\node[font=\fontsize{4}{4.5}\selectfont\mplfont, colOurs, inner sep=0.4pt, fill=white, fill opacity=0.75, text opacity=1, anchor=south, yshift=2.5pt] at (axis cs:41,54.45) {54.45};

\end{groupplot}

\node[font=\footnotesize, anchor=north]
    at ([yshift=-5mm]$(group c1r1.south)!0.5!(group c2r1.south)$) {Number of training videos};

\matrix[
    anchor=north,
    matrix of nodes,
    column sep=1.2mm,
    font=\scriptsize,
    every node/.style={inner sep=0pt, anchor=west},
] at ([yshift=-10mm]$(group c1r1.south)!0.5!(group c2r1.south)$) {
    \fill (0.12,0) circle (1.7pt); & \node{mAP}; &
    \node[draw, fill, rectangle, minimum size=3.2pt, inner sep=0pt] at (0.12,0) {}; & \node{HR@1}; &
    \node[draw, fill, regular polygon, regular polygon sides=3, minimum size=5pt, inner sep=0pt] at (0.12,0) {}; & \node{AUROC}; &
    \node[draw, fill, diamond, minimum size=4.5pt, inner sep=0pt] at (0.12,0) {}; & \node{AUPR}; &
    \draw[colOurs, line width=1.2pt] (0,0) -- (0.35,0); \fill[colOurs] (0.175,0) circle (1.5pt); & \node{Ours}; &
    \draw[colBaseline, line width=1pt, dashed] (0,0) -- (0.35,0); & \node{Baseline}; \\
};

\end{tikzpicture}
    \caption{
    Scaling of annotation-free training with the number of training videos (nested subsets; shaded band: $\pm$ std over five subsets per size).
    Dashed lines: NTSSL~\cite{parolari2026contrastive}.
    The break-even point is reached between 24 and 32 videos, depending on the metric, and performance keeps improving up to 41 videos.
    }
    \label{fig:exp-scaling}
\end{figure}

We next ask how much unlabeled full-procedure video is needed to offset the removal of spatial supervision.
We train on nested subsets of the 41-video training split (4, 8, 16, 24, 32, and 41 videos), with five independent subsets per size to average out the choice of videos.
Fig.~\ref{fig:exp-scaling} compares the resulting models with NTSSL~\cite{parolari2026contrastive}, which uses ground-truth boxes on the full training set.

Performance grows steadily with the number of videos: the annotation-free model matches NTSSL at around 24 videos on retrieval and AUPR, and at 32 videos on AUROC, and keeps improving up to the full 41-video set.
In line with scaling results on natural images~\cite{DBLP:conf/cvpr/Zhai0HB22}, more unlabeled videos can thus be more effective than fewer spatially annotated polyp observations.
This break-even point is specific to \rc{} and should not be read as a universal threshold.

\subsection{Robustness to Cohort Removal}
\label{sec:cohort}

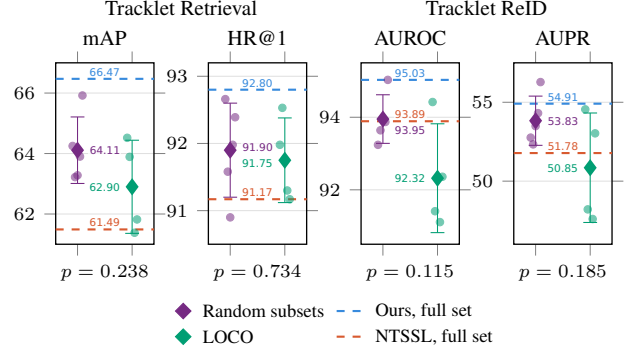
\begin{figure}[t]
\centering
\begin{tikzpicture}[font=\footnotesize,
    val/.style={font=\fontsize{4}{4.5}\selectfont\mplfont, inner sep=0.4pt, fill=white,
        fill opacity=0.75, text opacity=1},
]

\begin{groupplot}[
    group style={
        group size=4 by 1,
        horizontal sep=0.72cm,
    },
    scale only axis,
    width=1.3cm,
    height=2.4cm,
    xmin=-0.4, xmax=1.4,
    xtick={0,1},
    xticklabels={},
    ytick align=outside,
    xtick align=outside,
    tick label style={font=\scriptsize},
    title style={font=\footnotesize, yshift=-1mm},
    ymajorgrids, grid style={line width=0.2pt, draw=gray!25},
    axis line style={line width=0.5pt},
    clip=false,
]

\nextgroupplot[title={mAP}, ymin=61, ymax=67]
\draw[colOursFull, thick, dashed] (axis cs:-0.4,66.47) -- (axis cs:1.4,66.47);
\node[val, text=colOursFull, anchor=south, yshift=0.5pt] at (axis cs:0.5,66.47) {66.47};
\draw[colNtssl, thick, dashed] (axis cs:-0.4,61.49) -- (axis cs:1.4,61.49);
\node[val, text=colNtssl, anchor=south, yshift=0.5pt] at (axis cs:0.5,61.49) {61.49};
\addplot[only marks, mark=*, mark size=1.4pt, colRandom, opacity=0.55]
    coordinates {(-0.09,64.25) (-0.045,63.22) (0,63.28) (0.045,63.90) (0.09,65.92)};
\addplot[only marks, mark=*, mark size=1.4pt, colLoco, opacity=0.55]
    coordinates {(0.91,64.52) (0.955,63.89) (1.045,61.38) (1.09,61.82)};
\addplot[colRandom, mark=diamond*, mark size=2.6pt, thick,
    error bars/.cd, y dir=both, y explicit]
    coordinates {(0,64.11) +- (0,1.10)};
\node[val, text=colRandom, anchor=west, xshift=4pt] at (axis cs:0,64.11) {64.11};
\addplot[colLoco, mark=diamond*, mark size=2.6pt, thick,
    error bars/.cd, y dir=both, y explicit]
    coordinates {(1,62.90) +- (0,1.54)};
\node[val, text=colLoco, anchor=east, xshift=-4pt] at (axis cs:1,62.90) {62.90};

\nextgroupplot[title={HR@1}, ymin=90.5, ymax=93.2]
\draw[colOursFull, thick, dashed] (axis cs:-0.4,92.80) -- (axis cs:1.4,92.80);
\node[val, text=colOursFull, anchor=south, yshift=0.5pt] at (axis cs:0.5,92.80) {92.80};
\draw[colNtssl, thick, dashed] (axis cs:-0.4,91.17) -- (axis cs:1.4,91.17);
\node[val, text=colNtssl, anchor=south, yshift=0.5pt] at (axis cs:0.5,91.17) {91.17};
\addplot[only marks, mark=*, mark size=1.4pt, colRandom, opacity=0.55]
    coordinates {(-0.09,92.66) (-0.045,91.58) (0,90.90) (0.045,91.98) (0.09,92.39)};
\addplot[only marks, mark=*, mark size=1.4pt, colLoco, opacity=0.55]
    coordinates {(0.91,91.98) (0.955,92.53) (1.045,91.30) (1.09,91.17)};
\addplot[colRandom, mark=diamond*, mark size=2.6pt, thick,
    error bars/.cd, y dir=both, y explicit]
    coordinates {(0,91.90) +- (0,0.70)};
\node[val, text=colRandom, anchor=west, xshift=4pt, yshift=1pt] at (axis cs:0,91.90) {91.90};
\addplot[colLoco, mark=diamond*, mark size=2.6pt, thick,
    error bars/.cd, y dir=both, y explicit]
    coordinates {(1,91.75) +- (0,0.63)};
\node[val, text=colLoco, anchor=east, xshift=-4pt, yshift=-1pt] at (axis cs:1,91.75) {91.75};

\nextgroupplot[title={AUROC}, ymin=90.5, ymax=95.5]
\draw[colOursFull, thick, dashed] (axis cs:-0.4,95.03) -- (axis cs:1.4,95.03);
\node[val, text=colOursFull, anchor=south, yshift=0.5pt] at (axis cs:0.5,95.03) {95.03};
\draw[colNtssl, thick, dashed] (axis cs:-0.4,93.89) -- (axis cs:1.4,93.89);
\node[val, text=colNtssl, anchor=south, yshift=0.5pt] at (axis cs:0.5,93.89) {93.89};
\addplot[only marks, mark=*, mark size=1.4pt, colRandom, opacity=0.55]
    coordinates {(-0.09,93.24) (-0.045,93.64) (0,93.94) (0.045,93.87) (0.09,95.03)};
\addplot[only marks, mark=*, mark size=1.4pt, colLoco, opacity=0.55]
    coordinates {(0.91,94.42) (0.955,91.41) (1.045,91.11) (1.09,92.36)};
\addplot[colRandom, mark=diamond*, mark size=2.6pt, thick,
    error bars/.cd, y dir=both, y explicit]
    coordinates {(0,93.95) +- (0,0.67)};
\node[val, text=colRandom, anchor=north, yshift=-2pt] at (axis cs:0.5,93.95) {93.95};
\addplot[colLoco, mark=diamond*, mark size=2.6pt, thick,
    error bars/.cd, y dir=both, y explicit]
    coordinates {(1,92.32) +- (0,1.50)};
\node[val, text=colLoco, anchor=east, xshift=-4pt] at (axis cs:1,92.32) {92.32};

\nextgroupplot[title={AUPR}, ymin=46, ymax=57.5]
\draw[colOursFull, thick, dashed] (axis cs:-0.4,54.91) -- (axis cs:1.4,54.91);
\node[val, text=colOursFull, anchor=south, yshift=0.5pt] at (axis cs:0.5,54.91) {54.91};
\draw[colNtssl, thick, dashed] (axis cs:-0.4,51.78) -- (axis cs:1.4,51.78);
\node[val, text=colNtssl, anchor=south, yshift=0.5pt] at (axis cs:0.5,51.78) {51.78};
\addplot[only marks, mark=*, mark size=1.4pt, colRandom, opacity=0.55]
    coordinates {(-0.09,52.76) (-0.045,52.33) (0,53.49) (0.045,54.31) (0.09,56.28)};
\addplot[only marks, mark=*, mark size=1.4pt, colLoco, opacity=0.55]
    coordinates {(0.91,54.55) (0.955,48.21) (1.045,47.60) (1.09,53.03)};
\addplot[colRandom, mark=diamond*, mark size=2.6pt, thick,
    error bars/.cd, y dir=both, y explicit]
    coordinates {(0,53.83) +- (0,1.56)};
\node[val, text=colRandom, anchor=west, xshift=4pt] at (axis cs:0,53.83) {53.83};
\addplot[colLoco, mark=diamond*, mark size=2.6pt, thick,
    error bars/.cd, y dir=both, y explicit]
    coordinates {(1,50.85) +- (0,3.47)};
\node[val, text=colLoco, anchor=east, xshift=-4pt] at (axis cs:1,50.85) {50.85};

\end{groupplot}

\node[font=\footnotesize, anchor=south]
    at ([yshift=5mm]$(group c1r1.north)!0.5!(group c2r1.north)$) {Tracklet Retrieval};
\node[font=\footnotesize, anchor=south]
    at ([yshift=5mm]$(group c3r1.north)!0.5!(group c4r1.north)$) {Tracklet ReID};

\foreach \c/\p in {1/0.238, 2/0.734, 3/0.115, 4/0.185}
    \node[font=\scriptsize, anchor=north] at ([yshift=-1.5mm]group c\c r1.south) {$p = \p$};

\matrix[
    anchor=north,
    matrix of nodes,
    column sep=1.5mm,
    row sep=1.2mm,
    font=\scriptsize,
    every node/.style={inner sep=0pt, anchor=west},
] at ([yshift=-6.5mm]$(group c2r1.south)!0.5!(group c3r1.south)$) {
    \node[diamond, fill=colRandom, draw=colRandom, minimum width=6pt, minimum height=6pt, inner sep=0pt] at (0.175,0) {}; & \node{Random subsets}; &
    \draw[colOursFull, thick, dashed] (0,0) -- (0.35,0); & \node{Ours, full set}; \\
    \node[diamond, fill=colLoco, draw=colLoco, minimum width=6pt, minimum height=6pt, inner sep=0pt] at (0.175,0) {}; & \node{LOCO}; &
    \draw[colNtssl, thick, dashed] (0,0) -- (0.35,0); & \node{NTSSL, full set}; \\
};

\end{tikzpicture}
\caption{
Robustness to cohort removal: random subsets (five 32-video subsets drawn from all cohorts) against leave-one-cohort-out (LOCO) training.
Differences between the two protocols are not significant (Welch's $t$-test, $p$-values below each panel), indicating no large effect of removing a cohort.
}
\label{fig:cohort}
\end{figure}

We assess robustness to unseen acquisition domains by retraining our model from scratch with each of the four \rc{} cohorts withheld in turn (leave-one-cohort-out, LOCO) and evaluating it on the unchanged test set.
As a matched control, we use the 32-video random subsets of Sec.~\ref{sec:exp_scaling}, which remove a comparable amount of data drawn from all cohorts.
In Fig.~\ref{fig:cohort}, LOCO averages 62.90 mAP, above NTSSL (61.49), which uses ground-truth boxes and every cohort, and close to the control (64.11).
Re-identification is slightly more sensitive (92.32 against 93.95 AUROC for the control).
Withholding a cohort thus costs about as much as removing the same amount of data, and in mAP it still surpasses NTSSL trained on all cohorts, suggesting resilience to cohort removal.

\subsection{Ablation Study}

\begin{table}[t]
\caption{
Ablation toward annotation-free training.
\emph{Annot.}: polyp identities and boxes (\emph{ID+Box}), boxes (\emph{Box}), box statistics (\emph{Stats}), or none (--);
\emph{Loss}: supervised contrastive (\emph{SupCon}) or noise-aware (\emph{Noise-Aw.});
\emph{Views}: ground-truth box crops (\emph{BBox}), synthetic box crops (\emph{Synth.}), full frames (\emph{Frame}), or random crops (\emph{Crop});
\emph{Full}: full-procedure training.
Gray: annotation-dependent; \textbf{bold}: best annotation-free.
}
\label{tab:ablation}
\centering
\footnotesize
\setlength{\tabcolsep}{2.5pt}
\begin{tabular}{llcc|cc|cc}
\toprule

\multicolumn{4}{c|}{Training setup}
& \multicolumn{2}{c|}{Tracklet Retrieval}
& \multicolumn{2}{c}{Tracklet ReID}
\\

Annot. & Loss & Views & Full
& mAP & HR@1
& AUROC & AUPR
\\

\midrule
\rowcolor{gray!20}
ID+Box & SupCon & BBox & \xmark
& 57.52 & 84.24
& 94.33 & 43.81
\\

\rowcolor{gray!20}
Box & Noise-Aw. & BBox & \xmark
& 61.49 & 91.17
& 93.89 & 51.78
\\

\rowcolor{gray!20}
Stats & Noise-Aw. & Synth. & \cmark
& 65.60 & 91.71
& 95.14 & 57.72
\\

\midrule
-- & Noise-Aw. & Frame & \cmark
& 52.27 & 84.10
& 89.24 & 43.74
\\

-- & SupCon & Crop & \cmark
& 64.31 & 89.54
& 94.97 & 49.83
\\

-- & Noise-Aw. & Crop & \cmark
& \textbf{66.47} & \textbf{92.80}
& \textbf{95.03} & \textbf{54.91}
\\

\bottomrule
\end{tabular}
\end{table}

Tab.~\ref{tab:ablation} traces the path from polyp-centric supervised learning to our annotation-free setting.
Replacing identity labels with temporal self-supervision and the noise-aware loss raises retrieval mAP from 57.52 to 61.49 and ReID AUPR from 43.81 to 51.78, showing that temporal associations are an effective supervisory signal when their noise is addressed.
Switching to random crops on full-procedure videos with the same loss improves every metric further, reaching 66.47 mAP and 54.91 AUPR.
A supervised contrastive loss over the same temporal bags remains competitive (64.31 mAP), so the temporal sampling itself provides useful structure, but it lowers HR@1 and AUPR.
Full frames perform worst on every metric, while synthetic boxes, which rely on annotation statistics, trade lower retrieval scores for higher ReID AUPR.

\section{Conclusion}

We presented an annotation-free framework that learns polyp tracklet representations from raw full-procedure colonoscopy videos.
Without bounding-box or identity annotations, it achieves \sota{} results on \rc{} for tracklet retrieval, re-identification, and few-shot polyp retrieval, using only a modest number of unlabeled procedures.
Validation is currently confined to \rc{}: although it spans six centers on three continents and two manufacturers, it is the only public full-procedure dataset with persistent polyp identities, and we hope new datasets will enable broader validation.

\section{Declarations}
\label{sec:ethics}

\noindent\textbf{Ethics.}
This retrospective study uses open-access human subject data from \rc{}~\cite{biffi2024real}; ethical approval was not required, as confirmed by the data license.

\noindent\textbf{Competing interests.}
C.B. is affiliated with Cosmo Intelligent Medical Devices, developer of the GI Genius device.

{\spaceskip=.9\fontdimen2\font plus \fontdimen3\font minus 2\fontdimen4=-1
\noindent\textbf{Acknowledgments.}
We acknowledge ISCRA for awarding this project access to the LEONARDO
supercomputer, owned by the EuroHPC Joint Undertaking, hosted by CINECA
(Italy).\par}

\noindent\textbf{AI disclosure.}
Claude (Anthropic) was used to improve the writing and to prepare the charts; the authors reviewed and take responsibility for all content.

\bibliographystyle{IEEEbib}
\bibliography{strings,refs}

\end{document}